\documentclass[10pt,twocolumn,letterpaper]{article}

\usepackage[pagenumbers]{cvpr}

\usepackage{amsmath,amsthm}
\usepackage{multirow}
\usepackage{wrapfig}
\usepackage{graphicx}
\usepackage{svg}

\definecolor{cvprblue}{rgb}{0.21,0.49,0.74}

\newcommand{\needconf}[1]{\textcolor{black}{{#1}}}

\usepackage{dblfloatfix}

\usepackage[pagebackref,breaklinks,colorlinks,allcolors=cvprblue]{hyperref}
\usepackage[most]{tcolorbox}
\usepackage{xcolor,tcolorbox}
\definecolor{PromptOuterBlue}{RGB}{30,75,130}
\definecolor{PromptInnerTitle}{RGB}{235,245,255}
\definecolor{PromptInnerFrame}{gray}{0.7}

\def\paperID{14474} 
\def\confName{CVPR}
\def\confYear{2026}

\title{DiffGraph:  An Automated Agent-driven  Model Merging Framework for  \\ In-the-Wild Text-to-Image Generation}
\author{
Zhuoling Li$^{1}$ \quad Hossein Rahmani$^{1}$ \quad Jiarui Zhang$^{1}$ \quad Yu Xue$^{1}$ \\ \quad Majid Mirmehdi$^{2}$ \quad Jason Kuen$^{3}$ \quad Jiuxiang Gu$^{3}$   \quad Jun Liu$^{1}$\thanks{Corresponding Author}  
\\
$^1$Lancaster University \quad $^2$University of Bristol \quad $^3$Adobe Research\\
{\tt\small \{z.li81, h.rahmani, j.zhang76, y.xue9\}@lancaster.ac.uk,  m.mirmehdi@bristol.ac.uk,} \\
{\tt\small \{kuen, jigu\}@adobe.com, j.liu81@lancaster.ac.uk}
} 

\begin{document}
\maketitle
\begin{abstract}
The rapid growth of the text-to-image (T2I) community has fostered a thriving online ecosystem of expert models, which are variants of pretrained diffusion models specialized for diverse generative abilities. 
Yet, existing model merging methods remain limited in fully leveraging abundant online expert resources and still struggle to meet diverse in-the-wild user needs. We present DiffGraph, a novel agent-driven graph-based model merging framework, which automatically harnesses online experts and flexibly merges them for diverse user needs. Our DiffGraph constructs a scalable graph and organizes ever-expanding online experts within it through node registration and calibration. Then, DiffGraph dynamically activates specific subgraphs based on user needs, enabling flexible combinations of different experts to achieve user-desired generation. Extensive experiments show the efficacy of our method. 
Project page: \url{https://zhuoling.site/DiffGraph}.
\end{abstract}

\section{Introduction}
\label{sec:intro}
Benefiting from the extensive general knowledge acquired through large-scale pretraining, developers can fine-tune large diffusion models~\cite{Rombach_2022_CVPR,huggingfaceBlackforestlabsFLUX1devHugging} on their own customized datasets to produce specialized variants (e.g.,  checkpoints and LoRA-based models) endowed with task-specific generative capabilities, such as creating novel artistic styles~\cite{ruiz2023dreambooth, zhang2023inversion, gal2022image} or controlling particular visual attributes (e.g., image blur). We refer to these specialized variants as \textit{expert models}.
Nowadays, driven by the potential of text-to-image (T2I) generation, several online platforms (e.g., Civitai and Hugging Face) have rapidly flourished, where numerous experts with diverse skills are continuously developed, shared, and discussed, greatly enriching the online expert ecosystem and fostering a vibrant AI-driven creative community.

Motivated by the growing popularity of expert models, several recent studies~\cite{biggs2024diffusion, zhong2024multi,gu2023mix, wu2024mixture} have explored merging multiple experts to combine their specialized skills for more composite and complex image generation, such as generating an image of a specific character in a particular artistic style by merging a specific character expert and a style expert.
Despite the progress, existing approaches are mainly designed to merge a very small, fixed set of pre-provided experts~\cite{biggs2024diffusion, chen2024iteris, shah2024ziplora, wu2024mixture}, and often struggle to handle different (new) expert combinations without retraining~\cite{wu2024mixture} or test-time optimization~\cite{chen2024iteris}.
However, in-the-wild T2I generation scenarios are far more diverse. Users come from various backgrounds, such as graphic designers, web editors, and professional artists, and naturally exhibit distinct preferences and expectations for generated images. Consequently, prompts from different users, or even from the same user in different contexts, usually \textbf{require combining very different experts or even different numbers of experts} to satisfy highly varying in-the-wild user needs.

Meanwhile, online platforms host an abundant and ever-expanding ecosystem of experts, where new experts with novel generative skills are constantly emerging. 
Therefore, \textbf{fully organizing the large-scale online expert resources and flexibly utilizing (merging) them could unlock new opportunities to more effectively address diverse in-the-wild user demands.}
In terms of this, we note that some more recent methods~\cite{ouyang2025k, shenaj2024lora, li2025autolora} might have the potential to realize this vision due to their exploration of flexible merging of different experts.
A key commonality among these methods is that they primarily rely on using experts’ model parameters as input features.
For instance, LoRA.rar~\cite{shenaj2024lora} trains a supernetwork directly taking the parameter matrices of experts (typically one character LoRA and one style LoRA) as input features to predict merging coefficients for experts.
Despite their achieved efficacy, we argue that these parameter-dependent methods~\cite{ouyang2025k, shenaj2024lora, li2025autolora} are still not suitable for fully harnessing the real-world large-scale online resources. This is mainly because online experts, even when derived from the same pretrained diffusion models, can be developed using vastly different and even ever-emerging fine-tuning strategies~\cite{hu2022lora, liu2024dora, clarkdirectly}, diverse datasets, and varied training configurations.
As a result, \textit{online experts exhibit substantial and very complex parameter diversity and even architectural differences (e.g., checkpoint-based or LoRA-based experts)},
making these parameter-dependent model merging methods~\cite{ouyang2025k, li2025autolora, shenaj2024lora} struggle to generalize well to such diverse, heterogeneous, and time-evolving online resources. 
Moreover, these methods~\cite{ouyang2025k, shenaj2024lora, li2025autolora} are still limited to either combining a fixed number of experts or manually specifying the number of experts to be merged.

In this paper, we propose \textbf{DiffGraph},  a novel \textit{automated agent-driven graph-based model merging framework}, which automatically harnesses and flexibly merges various (and even various numbers of) online expert resources, serving for diverse user needs without any retraining or test-time optimization after deployment, while seamlessly scaling to the evolving expert ecosystem. 
Drawing on the insight that graphs can naturally encode
heterogeneous entities and their relationships~\cite{wu2022graph,
yang2021consisrec, feng2024graphrouter, hamilton2017inductive}, we
introduce a novel graph formulation tailored to expert management and merging.
As shown in~\cref{fig:main}, our DiffGraph incorporates two LLM-powered agents, namely, the \textit{Graph Construction Agent (GCA)}  and \textit{Expert Selection Agent (ESA)}  together with a variational graph autoencoder (VGAE)-based  \textit{Merging Planner (MP)}.
These components collectively provide the management and utilization of abundant online expert resources to fulfill in-the-wild T2I generation needs.

Specifically, \textbf{GCA} first constructs a \textit{universal graph} to manage online expert resources, by automatically collecting high-quality experts from public platforms and organizing them as \textit{expert nodes} in the graph through two complementary mechanisms: \textit{node registration} and \textit{node calibration}.
The \textit{node registration} mechanism qualitatively summarizes textual descriptions of experts' skills from their homepage information and encodes these descriptions as expert node features.
Meanwhile, the \textit{node calibration} mechanism quantitatively evaluates experts on a set of representative \textit{reference prompts} (i.e., \textit{reference prompt nodes} in the graph), and the resulting performance scores are represented as \textit{edge features} between the two types of nodes, as illustrated in~\cref{fig:main}.
On the other hand, the Expert Selection Agent (ESA) and Merging Planner (MP) operate on the universal graph constructed by GCA. When a user provides a prompt (referred to as a \textit{user prompt}), \textbf{ESA} parses the underlying requirements conveyed by the prompt and selects a set of experts with promising skills to fulfill these requirements. Then, \textbf{MP} activates a \textit{subgraph} centered on the selected expert nodes in the universal graph and temporarily inserts the user prompt into this subgraph as a \textit{user prompt node}. Within this subgraph, the node and edge features of the selected experts provide rich contextual cues, offering an informative and holistic characterization of their skills.
A trained VGAE is then employed to encode the contextual information, which then generates a high-quality merging scheme to combine the selected experts for generation.

Overall, our approach is effective and efficient. The construction of the universal graph is only to be performed once during the framework preparation, and can be stored locally in a lightweight format with minimal memory overhead. When online resources evolve, new experts can be conveniently inserted into the graph in a training-free manner through \textit{node registration} and \textit{node calibration} mechanisms. During inference,  our framework automatically and dynamically activates different subgraphs of the universal graph based on user needs, supporting flexible combinations of different experts and different numbers of experts.

Our contributions are as follows. 
1) To the best of our knowledge, this is the first attempt to fully autonomously collect, manage, and leverage large-scale online experts for model merging, which flexibly addresses diverse in-the-wild T2I generation needs.
2) By introducing the novel \textit{universal graph construction}  and \textit{dynamic subgraph activation} schemes, our framework can flexibly customize merging schemes based on user needs, while seamlessly scaling to newly emerging experts. To the best of our knowledge, this is also the first work investigating model merging from a novel \textit{graph-based} perspective.
3) Our method achieves state-of-the-art performance on evaluated benchmarks.

\section{Related Work}

\textbf{Diffusion Model Customization.}
Diffusion models~\cite{song2020denoising, dhariwal2021diffusion} have reshaped the field of AI-generated content~\cite{foo2025ai}.
With rich general knowledge acquired through large-scale pretraining, these models can be fine-tuned on customized datasets to derive various expert models with new generative capabilities~\cite{hui2025image,peng2024upam,zhang2024diff, gong2023diffpose, liu2025physics, li2025longdiff, liu2024floating, li2025diffip}. In addition to full-parameter fine-tuning for producing complete and powerful checkpoint-based experts, recent parameter-efficient fine-tuning (PEFT) techniques such as LoRA~\cite{hu2022lora} have gained popularity~\cite{zhu2024llafs,qu2024llms}, making model customization more accessible to individual developers.
Building on PEFT, developers can produce experts with only hundreds of training samples for more personalized and fine-grained creative control, such as generating unseen concepts~\cite{gal2022image,ruiz2023dreambooth} (e.g., new characters) or refining fine-grained visual attributes (e.g., enhancing facial or hand details). 
As a result of this customization surge, numerous high-quality experts are released and publicly available on online platforms.

\noindent
\textbf{Diffusion Model Merging.}  
Model merging~\cite{li2024selma,biggs2024diffusion,yu2024language,shah2024ziplora,chen2024iteris,shenaj2024lora,fengmodel,ouyang2025k,li2025autolora, gu2023mix, yang2024lora} aims to integrate multiple experts into a single model to combine their specialized skills. 
Some methods~\cite{wortsman2022model, li2024selma, biggs2024diffusion, fengmodel,yu2024language} focus on merging a small, fixed set of experts into a single model to enhance overall performance across diverse user (testing) prompts.
For instance, 
Diffusion Soup~\cite{biggs2024diffusion} introduces a greedy algorithm–based merging strategy to identify the optimal merging scheme.
Inspired by Particle Swarm Optimization~\cite{kennedy1995particle}, Model Swarms~\cite{fengmodel} treats each expert as a particle moving in the model parameter space, collaboratively searching for an optimal merged model.
Meanwhile, some other methods~\cite{shah2024ziplora, wu2024mixture, chen2024iteris} customize a merging scheme for each user prompt and often require retraining~\cite{chen2024iteris} or test-time optimization~\cite{wu2024mixture} to handle new expert combinations.
Besides, a few more recent methods~\cite{ouyang2025k, li2025autolora, shenaj2024lora} explore combining experts in a more flexible manner. For example, K-LoRA~\cite{ouyang2025k} conducts merging by comparing model parameter magnitudes, which enables different combinations of one character LoRA and one style LoRA in a training-free manner. AutoLoRA~\cite{li2025autolora} and LoRA.rar~\cite{shenaj2024lora} train networks that directly take the parameter matrices of experts as input to derive merged models.
Yet, these parameter-dependent methods~\cite{ouyang2025k, li2025autolora, shenaj2024lora} still struggle to generalize well to real-world online experts, due to the complex, heterogeneous, and diverse nature of the online resources.

Differently, our framework introduces a novel agent-driven, graph-based merging approach that dynamically organizes abundant online expert resources and adaptively combines different experts or even different numbers of experts to address diverse in-the-wild user needs.

\section{Proposed Method}

\begin{figure}[t] 
  \centering
 \includegraphics[width=0.95\linewidth]{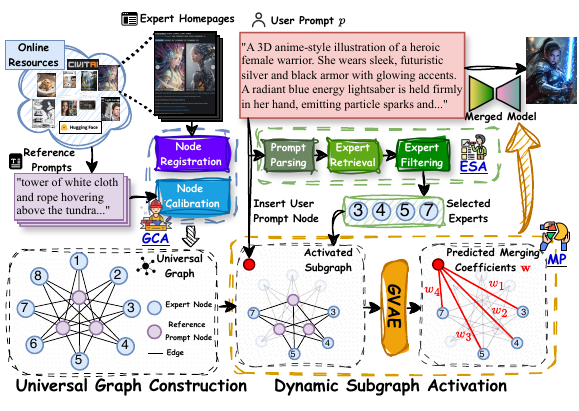}
\vspace{-0.45cm}
  \caption{
  Overview of our proposed method. Our DiffGraph framework consists of three key components: the Graph Construction Agent (GCA), the Expert Selection Agent (ESA), and the Merging Planner (MP). 
  In the \textit{Universal Graph Construction} stage, for ease of understanding, we illustrate a simplified example in which GCA collects and organizes \(N = 8\) online experts (indexed as 1--8), and evaluates them on \(N_r = 3\) reference prompts for node calibration.  During the \textit{Dynamic Subgraph Activation} stage, ESA parses user requirements and selects a subset of experts, for example \(\{3, 4, 5, 7\}\), to participate in the merging process. MP then activates the corresponding subgraph and generates the merging coefficients, which are used to produce the final merged model for image generation.
  }
  \label{fig:main}
\vspace{-0.45cm}
\end{figure}

\textbf{Overview.} Here, we give an overview of our proposed DiffGraph, the first automated agent-driven graph-based model merging framework, which comprehensively leverages online experts and flexibly customizes merging schemes to meet diverse in-the-wild image generation needs. 
As shown in~\cref{fig:main}, our framework comprises two LLM-powered agents, i.e., the Graph Construction Agent (GCA) and the Expert Selection Agent (ESA), as well as a VGAE-based module called the Merging Planner (MP). These three components collaboratively perform the following two key steps for expert merging:
\textbf{(1) Universal Graph Construction}: GCA organizes online expert resources into a graph (which we call the \textit{universal graph}) through novel node registration and calibration mechanisms (\cref{sec:construction}).
\textbf{(2) Dynamic Subgraph Activation}: Leveraging the strong reasoning capabilities of LLMs~\cite{qu2023lmc, li2024fewvs, zhang2026scenellm}, ESA autonomously and adaptively selects suitable experts capable of addressing user needs, and then MP activates the corresponding subgraph of the universal graph, upon which it customizes flexible merging schemes for image generation (\cref{sec:merge}).

\subsection{Universal Graph Construction}
\label{sec:construction}

To fully leverage the large-scale and ever-expanding online expert resources, our designed GCA first automatically scrapes online experts and then locally constructs the \textit{scalable universal graph}. Each collected expert is represented as a node (i.e., an \textit{expert node}), whose detailed features are subsequently initialized and enriched through two novel complementary mechanisms, namely \textit{node registration} and \textit{node calibration}. Since the efficacy of the universal graph depends on reliably capturing each expert’s specialized skills, these mechanisms are essential for facilitating the customization of merging schemes based on user needs.

\noindent\textbf{Node Registration.} 
Existing public platforms typically provide dedicated homepages for online experts, where developers present essential usage instructions and illustrative examples (e.g., applicative prompt examples). These materials offer useful insights into experts' capabilities. For each collected expert, we first initialize it in the universal graph as an isolated node, and then prompt an MLLM (e.g., GPT-4o) to generate a concise textual description of its skills based on the information available on its homepage (the detailed prompt is in \needconf{Supplementary}). 
The generated description is then encoded using a text embedding model (e.g., $\texttt{all-MiniLM-L6-v2}$~\cite{wang2020minilm}).
The resulting text embedding serves as the \textbf{node feature} for this expert.
It is worth noting that these text-embedding-based node features can also be used for efficiently pre-selecting experts from the large-scale online resources according to user needs, which will later participate \needconf{in the merging process (\cref{sec:merge}).}

\noindent\textbf{Node Calibration.}
Although the above node registration mechanism provides basic insights into expert skills, the resulting text-based node features may still be insufficient, offering only a qualitative understanding of experts' functionalities and failing to accurately reflect their actual image generation capabilities. 
To address this, we further design a \textit{node calibration} mechanism deriving edge features to characterize experts' skills from a quantitative perspective, offering cues for deriving high-quality merging schemes after the preliminary expert selection (detailed in~\cref{sec:merge}). 
Specifically, we construct a set of representative \textit{reference prompts} \needconf{$\{r_j\}_{j=1}^{N_r}$, where $N_r$} is the number of reference prompts. These prompts are added to the universal graph as a separate type of nodes, called \textit{reference prompt nodes}, and encoded using a text embedding model to serve as their node features. 
Then, each expert is evaluated on all reference prompts by feeding these prompts into the expert to generate corresponding images and then assessing their quality. The resulting quality scores are stored as \textbf{edge features} connecting that expert node to the corresponding reference prompt nodes, as illustrated in~\cref{fig:main}.
This process quantitatively characterizes each expert's capabilities. To achieve a comprehensive evaluation of generated images, we use multiple metrics 
(i.e., CLIP Score~\cite{hessel2021clipscore}, ImageReward~\cite{xu2023imagereward}, Aesthetic Score~\cite{schuhmann2022laion}, PickScore~\cite{kirstain2023pick}, and HPSv2~\cite{wu2023human}), concatenating the resulting evaluation scores to form the edge features. 
More details about the expert evaluation process are in \needconf{Supplementary}.
Moreover, to ensure that experts are calibrated on sufficiently diverse and representative prompts, we follow the prompt construction pipeline in~\cite{ye2024schedule} to enhance the linguistic and conceptual diversity of the reference prompt set. Specifically, we rank candidate prompts by analyzing key linguistic features such as nouns, prepositions, and adjectives, and select the most representative reference prompts. We provide more details about reference prompt construction in \needconf{Supplementary}.

Through the aforementioned \textit{node registration} and \textit{node calibration} mechanisms, GCA organizes online experts into the universal graph and establishes rich contextual information to characterize experts' capabilities. 
It is worth noting that the universal graph \textbf{can also seamlessly scale to the ever-evolving online expert ecosystem}. The GCA periodically monitors updates on public platforms. When a new expert (or a new version of an existing expert) emerges, GCA incorporates it into the universal graph by employing the \textit{node registration} and \textit{node calibration} mechanisms in a training-free manner. Conversely, when outdated experts are deprecated, their corresponding nodes can be simply removed from the graph, ensuring that the universal graph remains up to date. This whole process is very efficient. In our implementation, building a new universal graph containing \needconf{2319} online experts on a server with \needconf{4} A100 GPUs takes only \needconf{29} hours. Afterward, each new expert can be seamlessly incorporated into the graph within just \needconf{1.2} minutes. The constructed graph is lightweight and can be efficiently stored locally, with node connections saved in the \texttt{COO} format, and node as well as edge features stored as \texttt{.npz} files.

\subsection{Dynamic Subgraph Activation}
\label{sec:merge}
Below, we describe how our framework operates on the universal graph to customize merging schemes based on a given user prompt $p$.
A key insight underlying this process is that user prompts generally convey specific expectations and preferences that are usually relevant to only a subset of experts with corresponding skills. Therefore, involving all available experts in the merging process is unnecessary and also computationally intractable. In addition, current online experts are primarily released in two popular forms: \textit{checkpoint-based experts (abbreviated as CKPT experts)} and \textit{parameter-efficient fine-tuning-based experts (abbreviated as PEFT experts)}, such as LoRA~\cite{hu2022lora}.
As complete diffusion models, CKPT experts typically determine the main subjects and style characteristics of the generated images, shaping the overall generation quality.
Meanwhile, PEFT experts, which are built upon complete models, usually offer more fine-grained or personalized control over generation, such as refining artistic styles, adjusting lighting, or generating a specific anime character.
This functional distinction underscores the necessity of assigning CKPT and PEFT experts to distinct yet complementary roles, ensuring that each type effectively fulfills its intended function in the generation process. 

Considering the above, our LLM-powered Expert Selection Agent (ESA) first parses user needs and autonomously selects a set of experts from the universal graph that can collectively address these needs, which will later participate in the merging process.
Recall that the node registration mechanism (discussed in~\cref{sec:construction}) provides useful and qualitative insights into experts' capabilities. ESA leverages these insights to efficiently retrieve suitable experts from the large-scale available expert resources. 
Subsequently, the Merging Planner (MP) activates a specific subgraph of the universal graph centered on the selected expert nodes, and employs the VGAE to encode the contextual information established through the node registration and calibration mechanisms (discussed in~\cref{sec:construction}) within the subgraph, to derive an appropriate merging scheme.
By dynamically activating different subgraphs for different user prompts, our framework can flexibly aggregate experts that are needed by users while avoiding incorporating too many experts that may be irrelevant and degrade performance, thereby achieving user-desired image generation effectively and efficiently.
Below, we first describe the expert selection process, before detailing how the merging scheme is derived.

\noindent\textbf{Expert Selection.}
Our LLM-powered ESA first parses the user prompt \( p \) to filter out redundant information and generates a \textit{concise summary} \( s \) that captures basic image generation needs (e.g., the desired subject and overall artistic style). The summary \( s \) sets the tone for the desired image generation and serves as the key signal for selecting CKPT experts to offer necessary fundamental generation skills. Specifically, ESA retrieves from the universal graph the top-\( K_1 \) most relevant CKPT expert candidates, by encoding \( s \) into a text embedding and computing its similarity to experts' node features (i.e., the text embeddings of expert descriptions obtained in~\cref{sec:construction}). In addition to identifying basic generation needs for selecting CKPT experts, ESA then, in a chain-of-thought manner~\cite{wei2022chain}, decomposes the user prompt $p$ into a set of semantic components, such as the characters or scenes to generate, and then infers for each component the \textit{visual attributes} needed to depict it. 
For instance, to generate character portraits, it often needs to refine eye and lip details, while for scene elements, it usually involves adjusting lighting conditions.
In this way, ESA yields a set of fine-grained visual attributes \(\{a_m\}_{m=1}^{N_a}\). For each \( a_m \), ESA retrieves the top-\( K_2 \) most relevant PEFT expert candidates, by computing the similarity between the embedding of \( a_m \) and experts' node features.
The example of the LLM prompt used to generate the summary and extract the visual attributes is provided in \needconf{Supplementary}.

Ideally, one could directly prompt the LLM with the textual descriptions of all experts to select the most relevant ones. However, since there are thousands of experts available online, this approach is infeasible due to the LLM’s input token limitation. Therefore, we first employ the above text-embedding–based retrieval step to narrow down the search space, by filtering out clearly irrelevant candidates, obtaining $K_1$ CKPT expert candidates and  $N_a\times K_2$ PEFT expert candidates.
Yet, such a text-embedding–based retrieval alone may still include experts that are, in fact, not that relevant to the user's intent. For example, given the user prompt ``generate a portrait of a young woman with red hair", an expert described as ``specialized in generating red feathered flamingos" may also be retrieved. In addition, we empirically observe that experts with similar skills can be redundantly selected. Merging such irrelevant and redundant expert candidates can degrade generation quality and increase computational cost.

To address this issue, ESA performs an additional LLM-based filtering stage, where the LLM carefully reviews the textual descriptions of retrieved experts, and assesses their alignment with the parsed generation needs. This process finalizes the CKPT and PEFT experts to be merged, denoted as \(\mathcal{M}_\text{exp} = \{ \mathcal{M}_\text{ckpt}, \mathcal{M}_\text{peft} \}\), where \(\mathcal{M}_\text{ckpt} = \{ M_i^\text{ckpt} \}_{i=1}^{N_\text{ckpt}}\) is the selected CKPT experts, and \(\mathcal{M}_\text{peft} = \{ M_j^\text{peft} \}_{j=1}^{N_\text{peft}}\) denotes the selected PEFT experts. The numbers \(N_\text{ckpt}\) and \(N_\text{peft}\) are automatically determined by the LLM. More details of this filtering process are in  \needconf{Supplementary}.

\noindent\textbf{Merging Scheme Derivation.} 
With the selected experts $\mathcal{M}_\text{exp}$, we next describe how to effectively merge them to combine their expertise for achieving user-desired image generation. 
Specifically, the Merging Planner (MP) activates the nodes corresponding to the selected experts along with their directly connected one-hop neighboring nodes, forming a subgraph as shown in~\cref{fig:main}. Within this subgraph, the edge features together with the node features provide rich contextual information that characterizes each expert's capabilities. Moreover, the user prompt $p$ is attached to this subgraph as a temporary node, referred to as the \textit{user prompt node}, and the text embedding model is used to encode it as its node feature.
Then, a trained VGAE model is employed to encode the contextual information within the subgraph and generate the edge weights between the user prompt node and the selected expert nodes, as shown in~\cref{fig:main}. 
These edge weights serve as the \textit{merging coefficients}, indicating how much each expert contributes its specialized capability to the generation process. Finally, we obtain the merged diffusion model based on these coefficients and use it to generate the image for the user prompt $p$.
Below, we first define the activated subgraph and then elaborate on how the VGAE model generates the merging coefficients, based on which we produce the merged model.

\noindent\underline{The activated subgraph.}
As shown in~\cref{fig:main}, with the selected experts $\mathcal{M}_\text{exp} = \{ \mathcal{M}_\text{ckpt}, \mathcal{M}_\text{peft} \}$, MP activates a specific subgraph from the universal graph, denoted as $G=(\mathcal{V}, \mathcal{E}, \mathbf{X}, \mathbf{E})$. Here, $\mathcal{V} = \{v_p, \mathcal{V}_\text{ref},\mathcal{V}_\text{exp}\}$ denotes a node set consisting of the temporarily inserted user prompt node $v_p$, the reference prompt node set $\mathcal{V}_\text{ref}$, and the selected expert node set $\mathcal{V}_\text{exp}$; 
$\mathcal{E}$ denotes the set of edges between the selected experts and reference prompts;  
$\mathbf{X} = [\mathbf{x}_{p}^{\top}; \mathbf{X}_\text{ref}; \mathbf{X}_\text{exp}]\in \mathbb{R}^{|\mathcal{V}|\times d_{node}}$ denotes the node feature (i.e., text embedding with dimensionality $d_{node}$) matrix of $\mathcal{V}$, and $\mathbf{E}\in \mathbb{R}^{|\mathcal{E}|\times d_{edge}}$ is the edge feature with dimensionality $d_{edge}$, namely, the concatenation of the image quality scores (detailed in~\cref{sec:construction}) matrix of $\mathcal{E}$.

\noindent\underline{Merging coefficient generation.} 
Then, we employ a variational graph auto-encoder (VGAE)~\cite{kipf2016variational,zhao2024causality,zhang2025g} to encode the contextual information within the subgraph $G$ and predict the weights $\mathbf{w}\in\mathbb{R}^{|\mathcal{V}_\text{exp
}|}$ of edges between the user prompt node $v_p$ and the selected expert nodes $\mathcal{V}_\text{exp}$ as the merging coefficients. This process can be formulated as: 
\vspace{-0.1cm}
\begin{equation}
\label{eq:whole_process}
   \mathbf{w} = f(G; \theta) = Dec(\mathbf{w} \mid \mathbf{H}) Enc(\mathbf{H}\mid G),
\vspace{-0.1cm}
\end{equation}
where $f$ is the encoder-decoder-based VGAE parameterized by $\theta$, $Enc(\cdot)$ denotes the encoder module, and $Dec(\cdot)$ is the decoder module. $\mathbf{H}=[\mathbf{h}_p^{\top}; \mathbf{H}_\text{exp}]\in\mathbb{R}^{(1+|\mathcal{V}_\text{exp
}|)\times d_h }$ is concatenation of the latent vectors (with dimensionality $d_h$) of the user prompt node feature $\mathbf{x}_p$ and expert node features $\mathbf{X}_\text{exp}$, obtained by encoding the subgraph $G$ with the encoder $Enc(\cdot)$. This process can be further formulated as:
\vspace{-0.4cm}
\begin{equation}
\setlength{\abovedisplayskip}{3pt}
\setlength{\belowdisplayskip}{3pt}
\begin{aligned}
Enc(\mathbf{H} \mid G) &= \prod_{i=1}^{1+|\mathcal{V}_\text{exp}|} Enc(\mathbf{h}_{i} \mid G), \\
 \quad Enc(\mathbf{h}_i \mid G ) &= \mathcal{N}(\mathbf{h}_i \mid \boldsymbol{\mu}_i, \operatorname{diag}(\boldsymbol\sigma_i^2)),
\end{aligned}
\end{equation}
where $\boldsymbol{\mu} = \text{GNN}_{\boldsymbol{\mu}}(G; \theta_\mu)$ is the matrix of mean vectors $\boldsymbol{\mu}_i$, and  $\log(\boldsymbol{\sigma}) = \text{GNN}_{\boldsymbol{\sigma}}(G; \theta_\sigma)$ is the matrix of log-variance vectors $\boldsymbol{\sigma}_i$.
Here, $\mathbf{h}_i$, $\boldsymbol\mu_i$, and $\boldsymbol\sigma_i$ denote the $i$-th column of $\mathbf{H}$, $\boldsymbol{\mu}$, and $\boldsymbol{\sigma}$, respectively. 
We instantiate $\text{GNN}_{\boldsymbol{\mu}}$ and $\text{GNN}_{\boldsymbol{\sigma}}$ using simple two-layer GCNs, 
parameterized by $\theta_\mu$ and $\theta_\sigma$, respectively, and their architectural details are in \needconf{Supplementary}. 
The encoder $Enc(\cdot)$ is parameterized by $\theta_{enc} = \{\theta_\mu, \theta_\sigma\}$.

After encoding the contextual information within ${G}$ using the encoder $Enc(\cdot)$, the decoder $Dec(\cdot)$ then takes the resulting latent features $\mathbf{H} = [\mathbf{h}_p^{\top}; \mathbf{H}_\text{exp}]$ as input, and generates the edge weights $\mathbf{w} \in \mathbb{R}^{|\mathcal{V}_\text{exp}|}$ between the user prompt node $v_p$ and the selected expert nodes $\mathcal{V}_\text{exp}$. That is:
\vspace{-0.3cm}
\begin{equation}
Dec(\mathbf{w} \mid \mathbf{H}) = \prod_{i=1}^{|\mathcal{V_\text{exp}}|} Dec(w_{i} \mid \mathbf{h}_p, \mathbf{h}_{\text{exp},i};\theta_{dec}),
\vspace{-0.2cm}
\end{equation}
where \(\mathbf{h}_{\text{exp},i}\) denotes the \(i\)-th column of \(\mathbf{H}_\text{exp}\), \(\theta_{\text{dec}}\) represents the parameters of \(Dec(\cdot)\), and \(w_i \in [0,1]\) is the \(i\)-th element of \(\mathbf{w}\), serving as the merging coefficient for the \(i\)-th expert.
Considering that the reinforcement learning-based training described in~\cref{sec:train} requires a probabilistic prediction of $w_i$ rather than a deterministic value, we thus model the distribution of $w_i$ as a Beta distribution over $(0,1)$ with two parameters, $\alpha_i$ and $\beta_i$, predicted by a feed-forward network $\text{FFN}_{dec}$ parameterized by $\theta_{dec}$.
In addition, we note that ensuring $\alpha_i > 1$ and $\beta_i > 1$ results in a uni-mode Beta distribution, which is desirable as it prevents sampling potentially ambiguous or unstable $w_i$ from a multi-mode Beta distribution. 
To enforce this constraint, we re-parameterize $\text{FFN}_{dec}$ to predict two real-valued parameters, $a_i$ and $b_i$, namely, $ a_i, b_i = \text{FFN}_{dec}([\mathbf{h}_p, \mathbf{h}_{\text{exp},i}])$.
Then, $\alpha_i$ and $\beta_i$ are determined as $\alpha_i = 1 + e^{a_i}$ and $\beta_i = 1 + e^{b_i}$.
Consequently, the edge weight $w_{i}$
can be sampled as $w_i \sim \text{Beta}(\alpha_i, \beta_i)$.
Notably, during testing, we directly use the expectation of the predicted Beta distribution, as the deterministic merging coefficient for the $i$-th expert, i.e., $w_i =\frac{\alpha_i}{\alpha_i + \beta_i}$.

\noindent\underline{Merging Experts with $\mathbf{w}$.}  
Finally, we construct the merged model \(\mathbf{M}\) by combining the selected experts \(\mathcal{M}_{\text{exp}} = \{\mathcal{M}_\text{ckpt}, \mathcal{M}_\text{peft}\}\) according to the generated merging coefficients \(\mathbf{w}\). Specifically, we split \(\mathbf{w}\) into two parts, \(\mathbf{w}_\text{ckpt}\) and \(\mathbf{w}_\text{peft}\), corresponding to the CKPT and PEFT experts, respectively. The selected CKPT experts \(\mathcal{M}_\text{ckpt}\) are merged as \( W = \sum_{i=1}^{N_\text{ckpt}} \text{softmax}(\mathbf{w}_\text{ckpt})_i \cdot W_i \), where \( W_i \) denotes the model parameters of the \( i \)-th CKPT expert. Similarly, the selected PEFT experts \(\mathcal{M}_\text{peft}\) are merged as \( \Delta W = \sum_{j=1}^{N_\text{peft}} \text{softmax}(\mathbf{w}_\text{peft})_j \cdot \Delta W_j \), where \( \Delta W_j \) denotes the PEFT parameters of the \( j \)-th PEFT expert.
The two resulting components are then combined to obtain the parameters of the merged model  \( \mathbf{M} \), i.e., $\mathbf{W} = W +\Delta W$, where $\mathbf{W}$ is the parameters of $\mathbf{M}$.

Overall, the proposed \textit{Dynamic Subgraph Activation} is an effective and efficient mechanism for customizing merging schemes to handle diverse user prompts. By dynamically (selectively) activating the most promising experts for addressing user requirements, our framework tries to omit irrelevant and redundant experts that could degrade generation quality and increase computational overhead.

\subsection{Training and Testing}

\label{sec:train}
\noindent\textbf{Training.} 
The only learnable component in our framework is the lightweight VGAE model \( f(\cdot; \theta) \), where $\theta=\{\theta_{enc},\theta_{dec}\}$. At each training iteration, we randomly sample prompts from the training set (which do not overlap with the reference prompts) to simulate user prompts and then optimize VGAE to maximize the quality of generated images, formulated as $\arg\max_{\theta}\mathbb{E}_{\theta\sim\Omega}[u(I, p)]$, where $\mathbb{E[\cdot]}$ is the mathematical expectation, $\Omega$ is the parameter space,  $u(\cdot,\cdot)$ denotes the image-quality evaluation metric, and $I$ is the image generated using merging coefficients $\mathbf{w}$ for a user prompt $p$. 
Considering the excessively deep computational graph of the full denoising process, which hinders effective gradient backpropagation, we adopt policy gradient~\cite{williams1992simple, zhuge2024gptswarm} to approximate and optimize $\mathbb{E}_{\theta\sim\Omega}[u(I, p)]$:
\begin{equation}
\nabla_{\theta}\mathbb{E}_{\theta\sim\Omega}[u(I, p)]
\;\approx\;
\frac{1}{B} \sum_{b=1}^{B}u\!\left(I_{b}, p_b\right)\nabla_{\theta} P(\mathbf{w}_b),
\vspace{-0.1cm}
\end{equation}
where $B$ is the training batch size, $p_b$ is the $b$-th training prompt, $\mathbf{w}_b$ is the merging coefficients predicted by VGAE via~\cref{eq:whole_process}, $I_{b}$ is  
the corresponding generated image, and $P(\mathbf{w}_b)$ denotes the probability of $\mathbf{w}_b$ being sampled.
More details are in \needconf{Supplementary}.

\noindent\textbf{Testing.} 
Given a user prompt \( p \), ESA first parses the user requirements and selects a set of experts with promising skills to address these requirements, denoted as \(\mathcal{M}_\text{exp}\). 
Then, MP activates a subgraph $G$ centered around the selected expert nodes and employs the VGAE model to generate the merging coefficients \(\mathbf{w}\), which are subsequently used to construct the final merged model \(\mathbf{M}\) for image generation.

Notably, our framework can automatically organize abundant online experts, adaptively merge them to address user needs during testing, and effectively scale to new experts without fine-tuning. These advantages largely stem from our novel designs. Specifically, our node registration mechanism encodes experts' descriptive information into text-embedded node features to characterize their skills, which also enables efficient expert retrieval for a given user prompt via text-embedding similarity.
In addition, the proposed node calibration mechanism quantifies experts' capabilities by evaluating them on a set of reference prompts, producing performance score-based edge features that map all experts into a shared and meaningful capability space. Importantly, compared to the raw model parameters of experts, the text-encoded node features and  performance score-based edge features represent expert capabilities in a more interpretable and task-relevant manner and are thus more closely aligned with our ultimate objective, i.e., maximizing the quality of generated images. Hence, the VGAE, which takes these features as inputs, only needs to learn a relatively simple mapping from the shared capability space to merging coefficients, thereby enabling our framework to perform effective model merging.
Moreover, newly added experts can be seamlessly integrated by applying the same registration and calibration steps, allowing the pretrained VGAE to infer appropriate merging coefficients for them.

\section{Experiments}
\noindent
\textbf{Implementation Details.}
We conduct experiments on two widely used pretrained diffusion models: Stable Diffusion v1.5~\cite{Rombach_2022_CVPR} (abbreviated as SD15) and FLUX.1 Dev~\cite{huggingfaceBlackforestlabsFLUX1devHugging} (abbreviated as FLUX). For each pretrained diffusion model, high-quality derived expert models (released up to June 2025) are collected from popular public platforms, Civitai and Hugging Face, to construct the universal graph. 
We use GPT-4o to power the Graph Construction Agent and Expert Selection Agent,  and employ $\texttt{all-MiniLM-L6-v2}$~\cite{wang2020minilm} to encode text embeddings.
To train the VGAE model $f$ in the Merging Planner, we use the AdamW optimizer with beta coefficients $(0.9, 0.99)$ and an initial learning rate of $1\text{e-}2$.
More implementation details are in \needconf{Supplementary}.

\noindent\textbf{Datasets and Evaluation Metrics.}
To train the VGAE, we use DABench~\cite{zhao2024diffagent}, a dataset consisting of 50,482 in-the-wild user prompts collected from the online platform Civitai. These prompts feature complex combinations of visual attributes, reflecting the diverse and realistic needs of real users.  We evaluate our framework on the DABench test set and the widely used DiffusionDB~\cite{wang2023diffusiondb} benchmark, assessing image generation quality in terms of human preference and text faithfulness, reporting ImageReward (\textbf{IR})~\cite{xu2023imagereward}, HPSv2.1 (\textbf{HPS})~\cite{wu2023human}, PickScore (\textbf{PS})~\cite{kirstain2023pick}, Aesthetic Score (\textbf{AS})~\cite{schuhmann2022laion}, and CLIP Score (\textbf{CS})~\cite{hessel2021clipscore}.
More details on the dataset descriptions and splits are in \needconf{Supplementary}.

\begin{table}[t!]
\caption{Quantitative comparisons of different methods on T2I generation quality on the DABench and DiffusionDB datasets. \\$^*$ denotes methods equipped with our modified ESA module.}
\centering
\vspace{-0.35cm}
\scalebox{0.4}{
\renewcommand{\arraystretch}{1.2}
\begin{tabular}{l|l|ccccc|ccccc} 
\toprule
&\multirow{2}{*}{\textbf{Methods}} &  \multicolumn{5}{c}{\textbf{DABench}}  & \multicolumn{5}{c}{\textbf{DiffusionDB}} \\
& & \textbf{IR $\uparrow$} & \textbf{HPS $\uparrow$} &  \textbf{AS $\uparrow$}  & \textbf{PS $\uparrow$} & \textbf{CS $\uparrow$} & \textbf{IR $\uparrow$} & \textbf{HPS $\uparrow$}   &\textbf{AS $\uparrow$} &\textbf{PS $\uparrow$}  & \textbf{CS $\uparrow$} \\
\midrule
\multirow{10}{*}{SD15~\cite{Rombach_2022_CVPR}} 
    & Direct &-18.27&23.88&5.81&18.62&78.94     &14.83&23.74&5.87&19.61&82.70\\
    & DARE~\cite{yu2024language} &-3.86& 24.66 &5.84  &18.89 & 81.46     &28.02&24.78 &5.95&19.74&83.47 \\
    & Model Swarms~\cite{fengmodel}  &17.74&25.90&5.76&18.79 & 82.16   &50.62&26.63 &5.93&19.71&82.94\\
    & Diffusion Soup~\cite{biggs2024diffusion} &-3.81&25.55&5.92&19.41&81.70 &33.79&25.64&6.04&20.39  &84.61 \\
    & Ours fixed & \textbf{23.14}& \textbf{28.37}& \textbf{6.21}& \textbf{20.17}&\textbf{83.71} & \textbf{54.83}& \textbf{27.67}& \textbf{6.20}& \textbf{20.48}&\textbf{85.13}\\ 
    \cmidrule(lr){2-12}
    & ESA$^*$+K-LoRA~\cite{ouyang2025k} &19.33& 25.99& 5.97& 19.48 & 84.31 &27.14& 25.42&6.10 &19.94&85.16 \\
    & ESA$^*$+LoRA.rar~\cite{shenaj2024lora} &25.42&27.03&6.03&19.83&84.05 &34.23&25.46&6.14  &20.24&85.41 \\
    & AutoLoRA~\cite{li2025autolora}& 26.51& 27.41&6.04&19.96&82.97 &35.62 & 25.56&6.15 &20.17&83.02 \\
    & DiffAgent~\cite{zhao2024diffagent} &29.94 &27.83&6.36 &20.28 & 84.19 	 &52.65 &27.52&6.39& 20.31&84.84 \\
    & Ours & \textbf{73.11} &\textbf{30.06} & \textbf{6.54} &\textbf{20.62}&\textbf{84.79}	& \textbf{85.40}& \textbf{29.48}&\textbf{6.66} &\textbf{21.05}&\textbf{85.86} \\
\hline
\hline
\multirow{10}{*}{FLUX~\cite{huggingfaceBlackforestlabsFLUX1devHugging}} 
    & Direct &84.20&29.81&6.16&20.57&80.82	&91.14&29.40&6.13& 20.67&79.12\\
    & DARE~\cite{yu2024language} &95.74&29.60 &6.20&20.54&81.58 &100.21	&28.87 &6.18&20.87& 79.79  \\
    & Model Swarms~\cite{fengmodel} &	104.76	&30.03&6.27	&20.80&81.12		&120.82&30.22&6.17&21.26&81.07 \\
    & Diffusion Soup~\cite{biggs2024diffusion} 	&88.26&29.77&6.22&20.64 &80.63      &115.24&30.03 &6.10&21.25	&78.95\\
    & Ours fixed &\textbf{114.86}&\textbf{ 30.63}&\textbf{6.37}&\textbf{20.91}   &\textbf{81.73}   &\textbf{123.19}&\textbf{30.50}&\textbf{6.45}&\textbf{21.58}&\textbf{81.34}\\
    \cmidrule(lr){2-12}
    & ESA$^*$+K-LoRA~\cite{ouyang2025k} &88.22&30.36&6.12&20.23&83.38 &110.98&29.73&6.14&20.97&80.84\\
    & ESA$^*$+LoRA.rar~\cite{shenaj2024lora} 
    &86.15	&29.68&6.14&20.42&83.13&96.45&28.51&6.15&20.59&79.42\\
    & AutoLoRA~\cite{li2025autolora} &96.51&30.87&6.16&20.45&83.89&102.33&29.75&6.17&21.01&79.85\\
    & DiffAgent~\cite{zhao2024diffagent} &100.20&31.35&6.19&20.69&84.88&124.51&30.26&6.23&21.18&81.36\\
    & Ours &\textbf{136.62}&\textbf{32.72}	&\textbf{6.56}	&\textbf{21.25}&\textbf{85.02}&\textbf{148.75}&\textbf{32.36}&\textbf{6.64}&\textbf{21.84}&\textbf{82.66}\\
\bottomrule
\end{tabular}
}
\label{tab:general}
\vspace{-0.45cm}
\end{table}

\subsection{Evaluation on Text-to-Image Generation}
\label{sec:exp_general}

\noindent \textbf{Baseline Methods.}
\textbf{(1) Merging based on a fixed expert set.}
Some existing methods~\cite{fengmodel,biggs2024diffusion,yu2024language} operate on a small, fixed set of experts and combine all these experts into a single merged model to process user (testing) prompts. For a fair comparison with these methods~\cite{fengmodel,biggs2024diffusion,yu2024language}, we construct a fixed expert set by collecting the most popular experts across different category tags used by online platforms to categorize experts. This results in a set of \needconf{13} experts (details in \needconf{Supplementary}). To compare with these fixed expert set methods, we also develop a constrained variant (\textbf{Ours fixed}) of our framework by removing our ESA and directly using VGAE to derive merging schemes from the graph formed by all the experts in this fixed set.
We compare this variant with recent model merging approaches, including 
Diffusion Soup~\cite{biggs2024diffusion}, DARE~\cite{yu2024language}, and Model Swarms~\cite{fengmodel}. In addition, we include a simple baseline (\textbf{Direct}) that directly uses the original SD15 or FLUX for generation.
\textbf{(2) Leveraging Online Resources.} 
Besides the above approaches that focus on a small fixed expert set, we note that some recent methods~\cite{ouyang2025k, shenaj2024lora, li2025autolora} explore more flexible expert combinations and could have the potential to leverage large-scale online resources for model merging.
Among them, AutoLoRA~\cite{li2025autolora} retrieves and merges a prespecified number of experts. 
Meanwhile, K-LoRA~\cite{ouyang2025k} and LoRA.rar~\cite{shenaj2024lora} are designed to handle a fixed number of (yet different) experts 
during testing and can be extended to utilize online resources.
We adapt our ESA to automatically select two experts from online resources for each user prompt and feed them into these methods, enabling the use of online resources.
In addition, we also compare our method with DiffAgent~\cite{zhao2024diffagent}, which fine-tunes an LLM to find one suitable online expert for image generation.
Details of baselines and our variants are in \needconf{Supplementary}.

\begin{figure}[t]
    \centering
\includegraphics[width=0.9\linewidth]{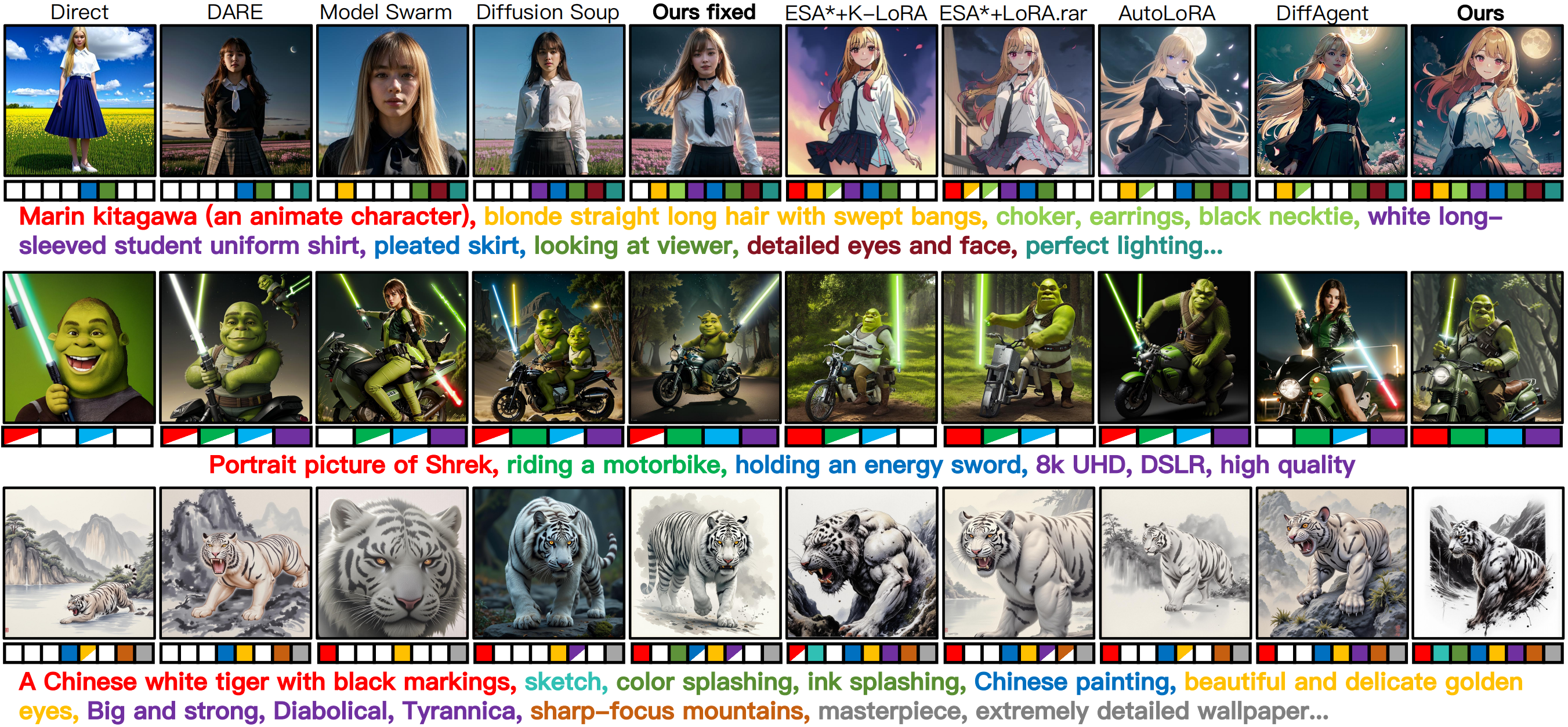}
    \vspace{-0.3cm}
     \caption{Qualitative comparisons of different methods on T2I generation.
     Different attributes in the prompt text are labeled with different colors.
     We illustrate visual attributes in a paint box manner, where a full colored cell denotes an attribute is successfully reflected in the generated image, a half colored cell denotes the attribute is reflected but at low quality, and an empty (white) cell means the corresponding attribute is totally missing. Zoom in for a better view. 
     More examples are in Supplementary.
     }
\vspace{-0.5cm}

 \label{fig:general_vis}
\end{figure}

\noindent \textbf{Results.} 
\cref{tab:general} shows the comparative results. Compared to the Direct baseline, methods like DARE, Diffusion Soup, and Model Swarms achieve better performance. 
\needconf{Notably, by performing model merging from the novel graph-based perspective,
our variant (Ours fixed) consistently outperforms all compared methods.}
On the other hand, we find that even when merging multiple online resources, K-LoRA, LoRA.rar, and AutoLoRA show limited improvements and even perform worse than the single-expert method, DiffAgent. A possible explanation is that these three methods mainly take the experts’ model parameters as input features to derive the merged model, which often struggle to handle large-scale online experts due to their substantial and complex model parameter diversity as well as architectural differences.
Differently, by representing experts through the proposed node registration and calibration mechanisms and deriving merging schemes from a novel graph viewpoint, our framework achieves the best performance.
Qualitative comparisons are shown in \cref{fig:general_vis}.

\subsection{Framework Analysis}
\label{sec:abalation}
Below, we use SD15 as the pretrained diffusion model to conduct experiments on DABench to further analyze our framework.  
More results are in Supplementary.
 
\setlength{\columnsep}{0.05in}
\begin{wraptable}[8]{r}{0.45\columnwidth}
\vspace{-0.45cm}
\caption{Evaluation on the scalability to new experts. 
}
\vspace{-0.35cm}
\centering
\resizebox{\linewidth}{!}{
\begin{tabular}{lcccccc} 
\toprule
 \textbf{Methods} &   \textbf{IR $\uparrow$} & \textbf{HPS $\uparrow$}& \textbf{AS $\uparrow$}  & \textbf{PS $\uparrow$} &\textbf{CS $\uparrow$}\\
\midrule     
    ESA$^*$+K-LoRA~\cite{ouyang2025k} 2023 &13.91	&25.88 &5.83&19.12 & 83.18\\
    ESA$^*$+K-LoRA~\cite{ouyang2025k} 2023$\rightarrow$2025 &  19.33 & 25.99 &  5.97& 19.48& 84.31 \\
    ESA$^*$+K-LoRA~\cite{ouyang2025k} 2025 &19.33& 25.99& 5.97& 19.48 & 84.31  \\
    ESA$^*$+LoRA.rar~\cite{shenaj2024lora} 2023 	&15.37&25.67&5.86&19.32 &83.45\\
    ESA$^*$+LoRA.rar~\cite{shenaj2024lora} 2023$\rightarrow$2025 &16.99 &26.08& 5.93 &19.39 &83.72\\
    ESA$^*$+LoRA.rar~\cite{shenaj2024lora} 2025 &25.42&27.03&6.03&19.83&84.05\\
    AutoLoRA~\cite{li2025autolora} 2023 &14.95&26.30&5.85	&18.91&81.43\\
    AutoLoRA~\cite{li2025autolora} 2023$\rightarrow$2025 &23.98&27.23&6.01&19.86&82.22\\
    AutoLoRA~\cite{li2025autolora} 2025 & 26.51& 27.41&6.04&19.96&82.97 \\
     DiffAgent~\cite{zhao2024diffagent} 2023 &26.33&27.37&6.30&20.12 &83.59\\
    DiffAgent~\cite{zhao2024diffagent} 2023$\rightarrow$2025 &26.33&27.37&6.30&20.12 &83.59\\
    DiffAgent~\cite{zhao2024diffagent} 2025 &29.94 &27.83&6.36 &20.28 & 84.19 \\
    \hline
    Ours 2023   &43.96&29.27&6.35&20.42&84.31\\
    Ours 2023$\rightarrow$2025 &69.64&29.66& 6.43&20.57&\textbf{84.81}\\
    Ours 2025 & \textbf{73.11} &\textbf{30.06} & \textbf{6.54} &\textbf{20.62}&84.79\\ 
\bottomrule
\end{tabular}
}
\label{tab:scale}
\end{wraptable}

\noindent\textbf{Scaling to New Experts.}
To evaluate whether our framework can scale to the ever-expanding online resources, we test it under the following three settings:
\textbf{(i) Ours 2023}: DiffGraph is trained using resources released up to June 2023, and resources released after this date are not used during evaluation.
\textbf{(ii) Ours 2023$\rightarrow$2025}: DiffGraph is trained using resources released up to June 2023, and new experts (released between June 2023 and June 2025) are incorporated into the framework via the node registration and calibration mechanisms in a \textit{training-free} manner (detailed in~\cref{sec:construction}), making this variant use resources released up to June 2025 during evaluation. 
\textbf{(iii) Ours 2025}: DiffGraph is trained using resources released up to June 2025 and utilizes all these resources during evaluation.
We also apply methods~\cite{ouyang2025k, shenaj2024lora, zhao2024diffagent, li2025autolora} to these settings for comparison.
Notably, among them, the LLM-based routing mechanism of DiffAgent~\cite{zhao2024diffagent} requires collecting user preference data for supervised training. Hence, it cannot leverage unseen (new) experts in a training-free manner, thus its results on the ``2023'' and ``2023$\rightarrow$2025''  settings are identical.
Besides, due to the training-free nature of K-LoRA~\cite{ouyang2025k}, its performance under the ``2023$\rightarrow$2025'' setting is the same as that under ``2025''.
As shown in~\cref{tab:scale}, our variant (Ours 2023$\rightarrow$2025) obtains results comparable to our full framework (Ours 2025) and even surpasses all other methods trained and evaluated on the ``2025" setting, showing our method scales effectively to the evolving expert ecosystem.

\setlength{\columnsep}{0.05in}
\begin{wraptable}[5]{r}{0.5\columnwidth}
\vspace{-0.5cm}
\caption{Evaluation on the mechanisms in graph construction.}
\vspace{-0.35cm}
\centering
\resizebox{\linewidth}{!}{
\begin{tabular}{lcccccc} 
\toprule
 \textbf{Methods} & \textbf{IR $\uparrow$} & \textbf{HPS $\uparrow$}& \textbf{AS $\uparrow$}  & \textbf{PS $\uparrow$} & \textbf{CS $\uparrow$} \\
\midrule     
     w/o registration &31.04&28.15&6.12&20.12&82.86\\
     Learnable  registration &33.36&28.67&6.20&20.21 &82.82\\
     w/o calibration  &11.92&25.87&5.94&19.56 &81.53\\
     Learnable calibration  &19.63&26.13&6.08&19.70& 81.02 \\
     Ours &\textbf{73.11}&\textbf{30.06}&\textbf{6.54} &\textbf{20.62}& \textbf{84.79}\\
\bottomrule
\end{tabular}
}
\vspace{-0.4cm}
\label{tab:initial}
\vspace{0.1cm}
\end{wraptable}
\noindent\textbf{Impact of Main Mechanisms in Universal Graph Construction.}
We verify the impact of the main mechanisms in universal graph construction by comparing the following variants:
\textbf{1) w/o registration}, where we remove the \textit{node registration} mechanism for constructing expert node features, and instead set them as zero embeddings;
\textbf{2) Learnable registration}, where we set the expert node features as learnable embeddings and optimize them during training;
\textbf{3) w/o calibration}, where we remove the \textit{node calibration} mechanism, namely, deleting all the edge features in the universal graph;
\textbf{4) Learnable calibration}, where we set the edge features as learnable embeddings and optimize them during training.
As shown in~\cref{tab:initial}, both the node registration and node calibration mechanisms contribute significantly to efficacy.

\setlength{\columnsep}{0.05in}
\begin{wraptable}[4]{r}{0.5\columnwidth}
\vspace{-0.45cm}
\caption{Evaluation on the Expert Selection Agent.}
\vspace{-0.38cm}
\centering
\resizebox{\linewidth}{!}{
\begin{tabular}{lcccccc} 
\toprule
 \textbf{Methods} & \textbf{IR $\uparrow$} & \textbf{HPS $\uparrow$}& \textbf{AS $\uparrow$}  & \textbf{PS $\uparrow$} & \textbf{CS $\uparrow$} \\
\midrule     
     w/o ESA &16.12&26.43&6.01&20.07&82.75\\ 
     
     Random activation &15.36&26.16&5.91&19.10&82.01\\ 
     Ours (full) &\textbf{73.11}&\textbf{30.06}&\textbf{6.54} &\textbf{20.62}& \textbf{84.79}\\
\bottomrule
\end{tabular}
}
\label{tab:ESA}
\end{wraptable}

\noindent\textbf{Impact of Expert Selection Agent.}
We explore the impact of ESA by comparing the following variants:
\textbf{1) w/o ESA}, where we remove ESA and directly employ MP to derive merging coefficients for all expert nodes in the universal graph, and select the top-10 (which we observe is the best performing number) experts with the highest merging coefficients for merging;
\textbf{2) Random activation}, where we remove ESA and, for each user prompt, randomly activate 10 expert nodes to form the subgraph.
The results in~\cref{tab:ESA} indicate  ESA  significantly contributes to performance.

\setlength{\columnsep}{0.05in}
\begin{wraptable}[5]{r}{0.5\columnwidth}
\vspace{-0.45cm}
\caption{Evaluation on the VGAE-based Merging Planner.}
\vspace{-0.35cm}
\centering
\resizebox{\linewidth}{!}{
\begin{tabular}{lcccccc} 
\toprule
 \textbf{Methods} & \textbf{IR $\uparrow$} & \textbf{HPS $\uparrow$}& \textbf{AS $\uparrow$}  & \textbf{PS $\uparrow$} & \textbf{CS $\uparrow$} \\
\midrule     
     w/o MP & 13.29& 26.20& 5.95& 19.73& 80.72\\
     Random merging &-5.27 &24.73&5.83 &19.0 &80.49 \\
     LLM-based merging&18.00 &25.87& 5.87 &19.62 &81.53\\
     Parameter-based merging&26.62& 27.24& 6.07 &20.00  &81.95\\
     Ours &\textbf{73.11}&\textbf{30.06}&\textbf{6.54} &\textbf{20.62}& \textbf{84.79}\\
\bottomrule
\end{tabular}
}
\label{tab:MP}
\end{wraptable}

\noindent\textbf{Impact of Merging Planner.}
Here, we investigate the impact of the VGAE-based MP by comparing the following variants:
\textbf{1) w/o MP}, where we remove VGAE and directly merge selected experts with equal weights;
\textbf{2) Random merging}, where we remove VGAE and merge  selected experts with randomly generated merging coefficients;
\textbf{3) LLM-based merging}, where we remove VGAE and prompt the LLM to produce the merging coefficients for  selected experts;
\textbf{4) Parameter-based merging}, where we follow existing parameter-dependent methods~\cite{li2025autolora, ouyang2025k, shenaj2024lora} and replace the node features of selected experts with their model parameters, which are then fed, together with the edge features, into VGAE to produce merging coefficients.
As shown in Tab.~\ref{tab:MP}, our method achieves better results than all other variants, showing the efficacy of MP that generates high-quality merging coefficients for image generation.

\section{Conclusion}  
We propose DiffGraph, a novel agent-driven graph-based model merging framework. Through the novel \textit{Universal Graph Construction} and \textit{Dynamic Subgraph Activation} schemes, our DiffGraph can effectively manage and utilize abundant online experts to meet diverse in-the-wild use needs. 
We achieve good results in our experiments.

{
    \small
    \bibliographystyle{ieeenat_fullname}
    \bibliography{main}
}

\end{document}